\documentclass{article}

    \PassOptionsToPackage{numbers, compress}{natbib}
\usepackage[final]{neurips_2022}

\usepackage[utf8]{inputenc} % allow utf-8 input
\usepackage[T1]{fontenc}    % use 8-bit T1 fonts
\usepackage{hyperref}       % hyperlinks
\usepackage{url}            % simple URL typesetting
\usepackage{booktabs}       % professional-quality tables
\usepackage{amsfonts}       % blackboard math symbols
\usepackage{nicefrac}       % compact symbols for 1/2, etc.
\usepackage{microtype}      % microtypography
\usepackage{xcolor}         % colors
\usepackage{algorithm}
\usepackage{algorithmic}
\usepackage{graphicx}
\usepackage{amsmath}
\usepackage{amsthm}
\usepackage{amsfonts}
\usepackage{booktabs}
\usepackage{algorithm}
\usepackage{algorithmic}
\usepackage{graphicx}
\usepackage{multirow}
\usepackage{tabularx}
\usepackage{booktabs}
\usepackage{url}
\usepackage{multirow}

\title{TIEG-Youpu's Solution for NeurIPS 2022 WikiKG90Mv2-LSC}

\author{%
  Feng Nie \\
  TIEG-Youpu\\
  Tencent, Shenzhen \\
  China\\
  \texttt{jannie@tencent.com} \\
  \And
  Zhixiu Ye \\
  TIEG-Youpu\\
  Tencent, Shenzhen \\
  China\\
  \texttt{zhixiuye@tencent.com} \\
  \And
  Sifa Xie \\
  TIEG-Youpu\\
  Tencent, Shenzhen \\
  China\\
  \texttt{sifaxie@tencent.com} \\
  \AND
  Shuang Wu \\
  TIEG-Youpu\\
  Tencent, Shenzhen \\
  China\\
  \texttt{seungwu@tencent.com} \\
  \And
  Xin Yuan \\
  TIEG-Youpu\\
  Tencent, Shenzhen \\
  China\\
  \texttt{bartyuan@tencent.com} \\
  \And
  Liang Yao \\
  TIEG-Youpu\\
  Tencent, Shenzhen \\
  China\\
  \texttt{dryao@tencent.com} \\
  \AND
  Jiazhen Peng \\
  TIEG-Youpu\\
  Tencent, Shenzhen \\
  China\\
  \texttt{brucejzpeng@tencent.com} \\
  \And
  Xu Cheng\thanks{Corresponding author.} \\
  TIEG-Youpu\\
  Tsinghua University, Beijing \\
  China\\
  \texttt{chengx19@mails.tsinghua.edu.cn} \\
}

\begin{document}

\maketitle

\begin{abstract}
  WikiKG90Mv2 in NeurIPS 2022 is a large encyclopedic knowledge graph. Embedding knowledge graphs into continuous vector spaces is important for many practical applications, such as knowledge acquisition, question answering, and recommendation systems. Compared to existing knowledge graphs, WikiKG90Mv2 is a large scale knowledge graph, which is composed of more than 90 millions of entities. Both efficiency and accuracy should be considered when building graph embedding models for knowledge graph at scale. 
  To this end, we follow the retrieve then re-rank pipeline, and make novel modifications in both retrieval and re-ranking stage. Specifically, we propose a priority infilling retrieval model to obtain candidates that are structurally and semantically similar. Then we propose an ensemble based re-ranking model with neighbor enhanced representations to produce final link prediction results among retrieved candidates. Experimental results show that our proposed method outperforms existing baseline methods and improves MRR of validation set from 0.2342 to 0.2839.  
\end{abstract}

\section{Introduction}
\begin{figure}[htbp]
\centering
\includegraphics[width=0.8\textwidth]{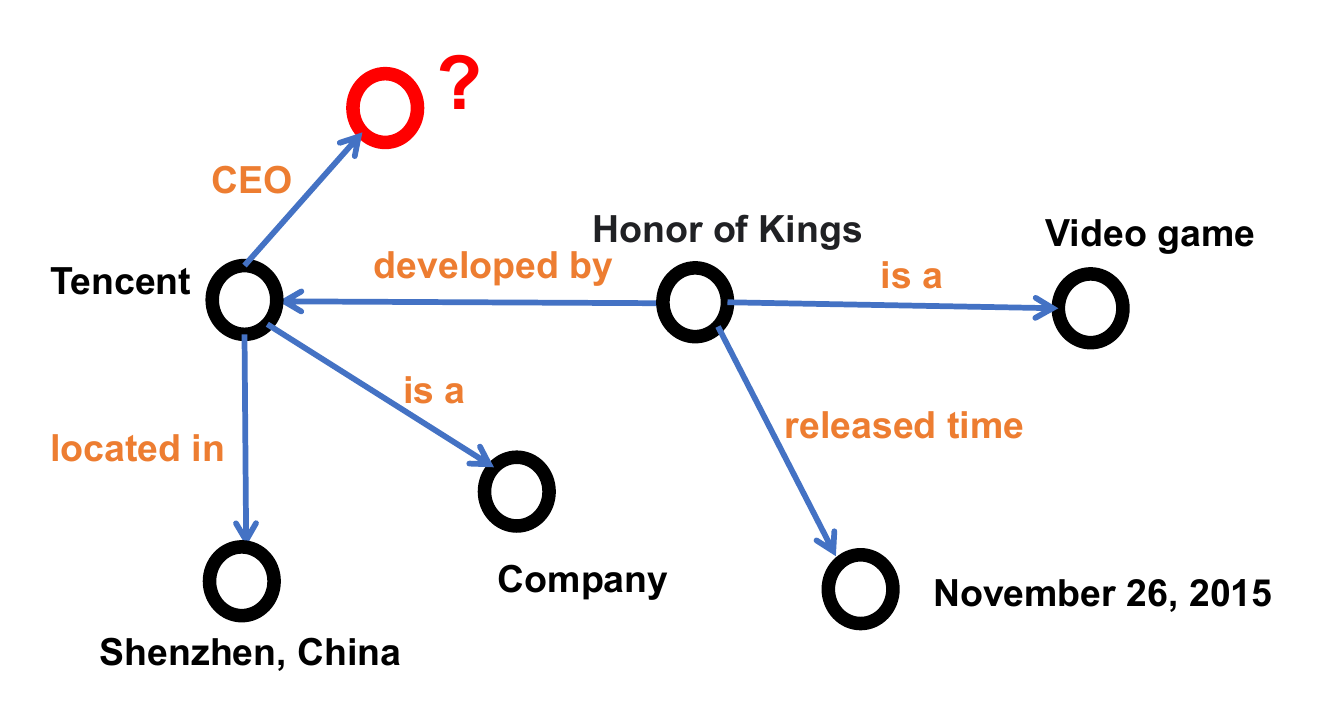}
\caption{An example of knowledge graph completion.}
\label{link_figure}
\end{figure}

Web-scale knowledge graphs (KGs) is important in both data mining and machine learning \cite{freebase,knowledge}, and plays an important role in various downstream applications such as question answering, knowledge acquisition, and recommendation systems. A typical knowledge graph is composed of entities and various relational edges, where each edge is represented as a triplet of the form $(head\ entity, relation, tail\ entity)$ ($(h, r, t)$ for short). Despite KGs contain rich structural information, they often suffer from knowledge incompleteness as world knowledge is updating rapidly\cite{knowledge}. Therefore, prediction over missing facts becomes a crucial task, also named as knowledge graph completion. Figure \ref{link_figure} depicts an example.

The 2022 NeurIPS releases the WikiKG90Mv2-LSC task, which focuses on correctly predicting missing facts in a large-scale KG. It is composed of more than 91 millions entities, a thousand relations, and 600 million triples. Directly building a complex graph embedding system and computing similarities with 91 millions of entities is time consuming. To this end, we follow the common strategy in large scale recommendation systems, by building a retrieval then re-ranking paradigm for link prediction. We make some novel modifications in both retrieval and re-ranking models. 
To retrieve most relevant candidates, we propose to leverage both structure and semantic information provided in knowledge graphs.  Specifically, we train multiple PIE models \cite{pie} with different strategies and additionally define 11 structural paths to fully leverage graph structure. Moreover, to incorporate semantic information, we directly use the original textual representations \cite{mpnet} provided by 2022 NeurIPS and apply k nearest neighbor search to generate semantically relevant entities. 
For re-ranking models, we train three types of knowledge graph embedding models TransE\cite{transe}, CompIEx \cite{complex}, NOTE\cite{ote,note} with different node presentations (i.e., randomly initialized embeddings, text feature embeddings, and graph-structure enhanced embeddings). Then we aggregate these results with a two step ensemble strategy. We conduct experiments on WikiKG90Mv2 dataset and improves MRR@10 of validation set from 0.2342 to 0.2839. Our team also achieves 0.2309 MRR in test-challenge dataset.

\section{Methodology}
Our proposed method is composed of two major components, a retrieval model and a re-ranking model. We will first introduce the retrieval model and then the re-ranking model in the following subsections.
\subsection{Retrieval Model}
Our retrieval model is composed of two steps. First, we design multiple retrieval models to exploit both structure or semantic information provided by KGs. Then, a priority infilling method is proposed to ensemble candidates retrieved by different retrieval models.
% We propose a structure enhanced retrieval and semantic embedding retrieval to leverage both structure and semantic information of KGs to retrieve as many relevant candidates as possible. Then, we propose a priority infilling method to ensemble candidates retrieved with different retrieval methods.
\subsection{Structural \& Semantic Enhanced Retrieval}
\noindent \textbf{PIE Retrieval}: we follow the official baseline method \cite{pie} and apply fine grained typing aware inference (PIE for short) to generate structurally similar candidates. This method utilizes neighborhood and relational types to search most relevant entities. Specifically, given an entity $e$ and a relation $r$, the candidates are selected based on the posterior distribution as follows 
$$p(e|r) = \frac{p(e)p(r|e)}{p(r)} \propto p(e)p(r|e), e \in N(e_q)$$
where $N(e_q)$ is the neighborhood entities of query entity $e_q$, $p(e)$ and $p(r)$ are prior distributions over entities and relations respectively (i.e., degrees). The entity typing model $p(r|e)$ is optimized with a self-supervised task by randomly masking several relations, and inference masked relations with remaining observed triplets in KG, more details can be find in \cite{pie}.

To retrieve diverse candidates, we train two PIE models with different size of sampled neighbors (we find $N\in\{6, 10\}$ achieves similar recall accuracy during experiments). Moreover, retrieval is more difficult for frequent relations, we therefore apply an up-sampling strategy by assigning higher weights for frequent relations samples. The weights are set accordingly where higher weights for relations with lower recall performance\footnote{Weights can be found in release code \url{https://github.com/CoderMusou/NeurIPS\_2022\_WikiKG90Mv2\_TIEG-Youpu}}.  

\noindent \textbf{Path Based Retrieval}: Despite PIE model is able to leverage graph structure information, we find that it performs poor in some infrequent relations. Therefore, we propose to directly leverage the structure information in KGs by defining several walking paths from entities or relations to entities.
%We reserve candidates for tail entities with highest walking probabilities. 
We follow the walking path settings in NOTE \cite{note}, and define 5 entity-to-entity paths (e.g., TH, HT, TH-HT, HT-HT, TH-HT) and 6 relation-to-entity paths (RT, RH, RT-HR-RT, RT-TR-RT, RH-HR-RT, RH-TR-RT). Take HT, TH-HT and RT-HR-RT as example, we define $F(h,r,t)$ to measure the co-occurrence possibility of a triplet $(h,r,t)$ as 
$$F_{HT}(h,r,t) = \frac{count(h,*,t)}{count(h,*,*)}$$

$$F_{TH-HT}(h,r,t) = \sum_{e_1} F_{TH}(e_1,*,h) \cdot F_{HT}(e_1,*,t) $$

$$F_{RT-HR-RT}(h,r,t) = \sum_{e_1,r_1} F_{RT}(*,r,e_1) \cdot F_{HR}(e_1,r_1,*) \cdot F_{RT}(*,r_1,t) $$
where $*$ denotes all possible entities or relations in KGs, $count(h,*,*)$ is the count of head entity $h$ in the whole OGB training dataset, and $F(h,*,t) = \sum_{r}F(h,r,t)$.
Finally, given $(h,r)$, we can retrieve 11 lists of tail entities sorted by $F(h,r,t)$ functions and we set 20,000 as the upper bound of the length of each list.

\noindent \textbf{Semantic Embedding Retrieval}: To incorporate semantic information of entities, we compute semantic similarity using text feature embedding produced by MPNet\cite{mpnet} as following 
$$F(h,r,t) = dist(\textbf{e}_r^{MPNet}, \textbf{e}_t^{MPNet})$$
where $dist(\cdot)$ denotes euclidean distance. We apply k nearest search with product quantization version of FIASS to enable fast retrieval. We only set k to 1000 according to validation set. 

\subsection{Priority Infilling Ensemble}
In order to ensemble results of recall methods mentioned above, we design a priority infilling ensemble method. First of all, we calculate the accuracy of each recall model $m$ as
$$accuracy(m) = \frac{ |S_{dev} \cap S_{m}| }{|S_{m}|}$$
where $S_{dev}$ is the set of $(h,r,t)$ in OGB valid set, $S_{m}$ is the set of all triplets $(h,r,t)$ retrieved by model $m$ and $|\cdot|$ is the size of a set. We then set the priority of each model according to $accuracy(m)$. Then we aggregate results of different models with priority from high to low until yielding $N$ candidates for each query $(h, r)$. $N$ is set to 20,000 according to validation set.
% and set a hyper-parameter $N$ as the upper bound of the number of candidates retrieved by the ensemble model. Finally, we infill the set of $N$ candidates for each $(h,r)$ in the order of the model priority.

\subsection{Re-ranking Model}
Our re-ranking model is composed of three steps. First, we propose a neighbor enhanced entity representation to aggregate first-order neighbor information directly for embedding initialization. Then, we apply several knowledge graph embedding methods to predict missing facts. Finally, we use an ensemble method to select most important graph embedding models for final predictions.

\noindent \textbf{Neighbor Enhanced Entity Representations}
Conventional choice for entity embedding initialization can be randomly initialized embeddings, and semantic embeddings (i.e., official text feature embedding with MPNet\cite{mpnet}). However, structure information is ignored with above methods. To address this issue, we propose a neighbor enhanced semantic embedding method to directly leverage graph structure into initialization stage. Specifically, each entity embedding $\textbf{e}_x^{struct}$ is represented by aggregating its first order neighbors entities as following.
\begin{equation}
   \textbf{e}_x^{struct} = \sum_{e_t \in N(e_x)}\textbf{e}_t^{MPNet}
\end{equation}
where $N(e_x)$ is the first order neighborhood entities of query entity $e_x$. In this way, three types of entity embedding are produced for graph embedding models to continue training.

\noindent \textbf{Graph Embedding Models}
For graph embedding models, we adopt advance algorithms in different domains to encode entities and relations, including TransE, NOTE and ComplEx. With these graph embedding methods, the model is capable of inference over various relations, such as 1-to-1, N-to-1, 1-to-N, reflective, inverse, symmetric and asymmetric relations. 

\noindent\textbf{TransE}: \citet{transe} interprets relation as a translation vector $\textbf{r}$, so that entities can be connected with simple translation, i.e., $\textbf{h} + \textbf{r} \approx \textbf{t}$. TransE is capable of capturing relation composition, but has difficulty in learning symmetric and asymmetric relations. 

\noindent \textbf{ComplEx}: \citet{complex} embeds entities and relations into complex domain. ComplEx is simple and efficient in capturing symmetric, and asymmetric relations by following: 
$$f(h,r,t) = Re(<\textbf{w}_r, \textbf{e}_h, \overline{\textbf{e}_t}>)$$
where $<\cdot>$ denotes hermitian product, $\textbf{e}_h = Re(\textbf{e}_h) + iIm(\textbf{e}_h)$ is a vector that contains both real vector components and imaginary vector components. $\overline{\textbf{e}_h} = Re(\textbf{e}_h) - iIm(\textbf{e}_h)$ refers to conjugate of vector $\textbf{e}_h$. $Re(\cdot)$ denotes taking the real vector component.

\noindent \textbf{NOTE}: is a normalized version of OTE model\cite{ote}. OTE models relations as group-based orthogonal transform embedding, which is able to model symmetric, inverse and compositional relations by simple transposing. The scoring function is defined as 
\begin{equation}
    f((h, r), t) = \sum_{i=1}^{K}(||\textbf{s}_r^{h}(i)\phi(\textbf{M}_r(i))\textbf{e}_h(i) - \textbf{e}_t(i)||) 
\end{equation}
\begin{equation}
     f(h, (r, t)) = \sum_{i=1}^{K}(||\textbf{s}_r^{t}(i)\phi(\textbf{M}_r(i))^T\textbf{e}_t(i) - \textbf{e}_h(i)||)
\end{equation}
where $\textbf{s}_r^h(i) = \frac{diag(exp(\textbf{s}_r(i))}{||diag(exp(\textbf{s}_r(i))||}$ and $\textbf{s}_r^t(i) = \frac{diag(exp(-\textbf{s}_r(i))}{||diag(exp(-\textbf{s}_r(i))||}$ are the weights of relation matrix, $\phi$ is the Gram-Schmidt process.

\noindent \textbf{Model Selection}
In this competition, we train TransE, ComplEx and NOTE with different entity embeddings (i.e., randomly initialized embeddings, text feature embeddings, graph structure enhanced embeddings) and hyper parameters. 
In order to combine results produced by multiple knowledge graph embedding models, we merge these results with a two-step ensemble strategy. First, we apply a greedy search to decide whether current model is qualified or not for final prediction. With the first filtering step, only 6 models are selected to produce the link prediction results. Then, we apply grid search to learn importance of each model.  
\section{Experiments}
\subsection{Experimental Details}
WikiKG90Mv2 dataset contains three time-stamps: May,17th, June 7th, and June 28th of 2021 for training, validation and testing respectively. We only use training dataset to train recall and re-ranking models, and select hyper parameters based on validation set. 
For PIE recall model, we use the following hyper parameters for model training, where batch size is 512, context hops is 3, learning rate is 2e-3, hidden dimension is 1024, margin and gamma are set to 3, and number of sampled neighbors is 10.

For graph embedding models, we use the following hyper parameters as default NOTE setting, where batch size is 1000, hidden dimension is 200, orthogonal vector size is 20, learning rate is 0.1, regularization coefficient is 1e-9, negative sample size is 1000, learning rate for entity encoder is 4e-5, learning rate decay step is 2000. 
For TransE and ComplEx, which are simpler than NOTE, we therefore set a larger batch size (16384), hidden dimension (600) and negative sampling size (16384). During experiments, we find that NOTE achieves better results using the combination of text feature/structure enhanced embedding and randomly initialized embeddings, while TransE and ComplEx are more suitable with randomly initialized embeddings. Note that we use 4 A100 GPUs to train each graph embedding model. 

\subsection{Experimental Results}
\begin{table}[]
\centering
\scalebox{0.74}{
\begin{tabular}{cccc}
\hline
\toprule
\textbf{Method}    & \textbf{Main Parameter Settings}                         & \textbf{Recall @20000} & \textbf{ Acc.} \\ \hline
%\multicolumn{4}{c}{\textbf{PIE Based Structural Retrieval Models}}                                                                                      \\ \hline
\multicolumn{4}{c}{\textbf{Structure Based Retrieval Models}}                                                                                      \\ \hline
PIE\_6             & neighbor samples  = 6                                    & 0.6008                             & 4.18e-5  \\
PIE\_10            & neighbor samples = 10                                    & 0.6044                             & 3.73e-5           \\
PIE\_10\_upsample  & neighbor samples = 10 w/ up sampling                    & 0.6044                    & 3.79e-5           \\ %\hline
%\multicolumn{4}{c}{\textbf{Rule Based Structural Retrieval Models}}                                                                                     \\ \hline
Rule-1                 & HT                                                        &0.0568                            & \textbf{5.46e-3}            \\
Rule-2                 & TH                                                      & 0.0488                            & 7.42e-4            \\
Rule-3                 & RT                                                       & 0.582\ \ \                              & 6.43e-5           \\
Rule-4                & RH                                                        & 0.0244                            & 1.66e-6           \\
Rule-5              & TH-TH                                                       & 0.0332                           & 8.76e-5               \\
Rule-6              & HT-HT                                                         & 0.1962                           & 5.50e-4           \\
Rule-7              & TH-HT                                                        & 0.0958                            & 5.11e-4            \\
Rule-8          &RT-TR-RT                                                        & \textbf{0.6229}                             & 3.32e-05           \\ 
Rule-9           & RT-HR-RT                                                       & 0.3308                             & 1.75e-05        \\
Rule-10           & RH-HR-RT                                                         & 0.5641                             & 2.99e-05           \\
Rule-11          & RH-TR-RT                                                        & 0.2126                             & 1.13e-05           \\

\hline
\multicolumn{4}{c}{\textbf{Semantic Based Retrieval Models}}                                                                                \\ \hline
MPNet Recall       & FIASS w/ product quantize centroids = 64, code size = 64 & 0.2394                             & 1.197e-05            \\ \hline
\multicolumn{4}{c}{\textbf{Ensemble Based Methods}}                                                                                       \\ \hline
direct structural ensemble    & majority voting                                          & 0.7013                            & -                    \\
structural priority infilling & ensemble structural results based on  model priority & 0.7136                   
& -                    \\
structural and semantic priority infilling & ensemble all results based on model priority & \textbf{0.7361}                    & -                    \\
\hline
\toprule
\end{tabular}
}
\caption{Recall results based with structural and semantic enhanced retrieval models}
\label{tab:retrieval}
\end{table}

\begin{table}[]
\centering
\scalebox{0.7}{
\begin{tabular}{ccc}
\hline
\toprule
\textbf{Method}                        & \textbf{Main Parameter Settings}                                                                                                                                     & \textbf{Validation MRR@10} \\ \hline
TransE-0                      & -                                                                                                                                                           & \textbf{0.214\ \ }    \\
TransE-1                      & batch size = 20480, negative sample size = 20480                                                                                                            & 0.2094            \\
TransE-2                      & \begin{tabular}[c]{@{}c@{}}batch size = 20480, negative sample size = 20480, \\ MPNet and randomly initialized embeddings w/ MLP layer\end{tabular}         & 0.2114            \\
TransE-3                      & neighbor enhanced embeddings and randomly initialized embeddings w/ MLP layer                                                                               & 0.1877            \\
ComplEx                       & -                                                                                                                                                           & 0.1649            \\
NOTE-0                         & -                                                                                                                                                           & 0.1561            \\
NOTE-1                         & negative sample size = 1200                                                                                                                                 & 0.1648            \\
NOTE-2                         & neighbor enhanced embeddings and randomly initialized embeddings w/ MLP layer                                                                               & 0.1654            \\
NOTE-3                         & \begin{tabular}[c]{@{}c@{}}negative sample size = 1200, \\ neighbor enhanced embeddings and randomly initialized embeddings w/ MLP layer\end{tabular}       & 0.1592            \\ \hline
Direct ensemble               & grid search above all models with non-zero weights                                                                                                                            & 0.28\ \ \ \               \\ \hline
Ensemble with Model Selection & \begin{tabular}[c]{@{}c@{}}first select best models \\ (TransE-0,TransE-1,TransE-2,ComplEx, OTE-0, OTE-2) \\ then grid search on model weights\end{tabular} & \textbf{0.2839}   \\ \hline
\toprule
\end{tabular}
}
\caption{Results of different graph embedding models using generated candidates on validation set.}
\label{tab:rerank}
\end{table}

\noindent \textbf{Retrieval Results:} Table \ref{tab:retrieval} reports retrieval results of different methods. Note that we only reserve 6 rule based recall models with higher accuracy than PIE's and limit the candidate size to 20,000 according to validation results. 
Table \ref{tab:retrieval} show that structural enhanced retrieval models generally achieves better result than semantic based retrieval models. Combination of results produced by PIE models and rule based models can lead to huge improvements. When incorporating semantic retrieval methods, the results can be improved further.  

\noindent \textbf{Re-ranking Results:} Table \ref{tab:rerank} presents the MRR@10 of different graph embedding methods based on generated candidates. TransE performs best compared to some state-of-art graph embedding models such as ComplEx and NOTE on WikiKG90Mv2 surprisingly. 
The result also shows the necessity of ensemble of these different graph embedding models. Combinations of these models leads to significant improvement, which indicates that different graph embedding models are good at prediction on different types of relations. Moreover, the results can be further improved with appropriate model selection strategy. 

\section{Conclusion}
In this paper, we present our solution for link prediction task on WikiKG90Mv2 dataset. Our proposed method follows the retrieval and re-ranking paradigm, and makes some novel modifications in both retrieval and re-ranking step. 
Specifically, for candidate retrieval, we propose to leverage structure and semantic information during retrieval to select relevant candidates. 
Moreover, we propose a priority infilling ensemble technique to merge candidate results produced by different retrieval models.  
For re-ranking step, we first enhance the original node representation by aggregating first order neighbors and then train multiple state-of-art graph embedding models including TransE, ComplEx and NOTE. Then we ensemble these results with a model selection strategy and grid search. The experimental results show effectiveness of our proposed method. In the future, we will consider improving the efficiency of candidate retrieval models.

\bibliographystyle{plainnat}
\bibliography{neurips_2022}

\end{document}